\documentclass[letterpaper, 10 pt, journal, twoside]{IEEEtran}
\ifCLASSINFOpdf
\else
\fi
\usepackage{graphics} 
\usepackage{epsfig} 
\usepackage{mathptmx} 
\usepackage{times} 
\usepackage{amsmath} 
\usepackage{amssymb}  
\usepackage{amsfonts}
\usepackage{xcolor}
\usepackage{url}
\usepackage{subcaption} 
\usepackage{booktabs}
\usepackage{cite}
\usepackage{multirow}
\usepackage{threeparttable}
\usepackage{orcidlink}

\begin{document}
%
\title{HybridTrack: A Hybrid Approach for Robust Multi-Object Tracking
}
%
%
%


\author{
Leandro Di Bella$^{1,2}$~\orcidlink{0009-0000-1731-7205}, 
Yangxintong Lyu$^{1,2}$~\orcidlink{0000-0002-2501-9010}, 
Bruno Cornelis$^{1}$~\orcidlink{0000-0002-0688-8173}, 
and Adrian Munteanu$^{1,2}$~\orcidlink{0000-0001-7290-0428}~\IEEEmembership{Member,~IEEE}%
\thanks{Manuscript received: December 30, 2024; Revised April 7, 2025; Accepted May 10, 2025.}
\thanks{This paper was recommended for publication by Editor Hyungpil Moon upon evaluation of the Associate Editor and Reviewers' comments.}
\thanks{This work was supported by Innoviris within the research project TORRES.}
\thanks{$^{1}$Leandro Di Bella, Yangxintong Lyu, Bruno Cornelis, and Adrian Munteanu are with the Department of Electronics and Informatics, Vrije Universiteit Brussel, Pleinlaan 2, B-1050 Brussels, Belgium.}
\thanks{$^{2}$Leandro Di Bella, Yangxintong Lyu, and Adrian Munteanu are also with IMEC, Kapeldreef 75, B-3001 Leuven, Belgium.}

\thanks{Digital Object Identifier (DOI): see top of this page.}
}

%
%

\markboth{IEEE Robotics and Automation Letters. Preprint Version. Accepted May, 2025}
{Di Bella \MakeLowercase{\textit{et al.}}: HybridTrack: A Hybrid Approach for Robust
Multi-Object Tracking} 

%



\maketitle

\begin{abstract}
The evolution of Advanced Driver Assistance Systems (ADAS) has increased the need for robust and generalizable algorithms for multi-object tracking. Traditional statistical model-based tracking methods rely on predefined motion models and assumptions about system noise distributions. Although computationally efficient, they often lack adaptability to varying traffic scenarios and require extensive manual design and parameter tuning. To address these issues, we propose a novel 3D multi-object tracking approach for vehicles, HybridTrack, which integrates a data-driven Kalman Filter (KF) within a tracking-by-detection paradigm. In particular, it learns the transition residual and Kalman gain directly from data, which eliminates the need for manual motion and stochastic parameter modeling. Validated on the real-world KITTI dataset, HybridTrack achieves 82.72\% HOTA accuracy, significantly outperforming state-of-the-art methods. We also evaluate our method under different configurations, achieving the fastest processing speed of 112 FPS. Consequently, HybridTrack eliminates the dependency on scene-specific designs while improving performance and maintaining real-time efficiency.\footnote{The code is publicly available at: \url{https://github.com/leandro-svg/HybridTrack}.}
\end{abstract}

\begin{IEEEkeywords}
Multi-Object Tracking, Kalman Filtering, Autonomous Driving\end{IEEEkeywords}

%
\IEEEpeerreviewmaketitle

\section{Introduction}
\label{sec:intro}
\IEEEPARstart{M}{ulti-object} tracking (MOT) involves locating and associating objects across consecutive image and/or LiDAR frames. It allows us to predict object trajectories, handle occlusions, and manage object appearance, making it widely used in Advanced Driver Assistance Systems (ADAS) to enhance traffic safety. The existing MOT techniques can be classified into various paradigms, including \textbf{Joint Detection and Tracking} (JDT) \cite{tokmakov2021learning}, \cite{tokmakov2022object},\cite{wu2021tracklet}, \cite{10777493} and \textbf{Tracking-by-Detection} (TBD) \cite{1211504}. JDT combines detection and tracking in a unified end-to-end deep-learning framework, leveraging temporal and spatial information for robust performance. 
However, the combination of detection and tracking reduces the modularity of the entire framework, increasing computational complexity. This complexity can limit applicability of the tracking algorithms in real-time traffic situations, particularly on resource-constrained systems such as smart cameras. Furthermore, the requirements for large-scale tracking data and the extensive training complexity of an end-to-end pipeline, make the use of JDT for real-world MOT potentially inefficient. 

{In contrast to JDT, TBD separates detection from tracking, enabling greater flexibility and computational efficiency. This approach allows state-of-the-art object detectors \cite{shi2020pv, wu2023virtual, wu2022casa} to be easily integrated with tracking algorithms. A notable strength of TBD is its use of recursive frameworks for the tracking component, such as Kalman \cite{welch1995introduction} or Bayesian filtering \cite{chen2003bayesian}, which continuously update the state of an object based on prior estimates and new observations. This recursive process enhances robustness and adaptability in dynamic environments. However, the reliance of state-of-the-art methods on these traditional model-based approaches presents a significant bottleneck. Specifically, the traditional methods rely on predefined motion models specific to traffic situations e.g., constant turn rate for cars, kinetic bicycle models for bicycles, and constant acceleration for pedestrians \cite{weng20203d}, \cite{wu20213d}, \cite{cao2023observation}, \cite{wang2023camo}. Such specificities limit adaptability to different traffic agents, degrading performance.}

{Moreover, estimating noise parameters such as process and measurement covariances is challenging with complex sensor data, i.e. LiDAR or image data, requiring extensive manual tuning for specific scenarios. For instance, covariances optimized for highways fail in dense urban settings with frequent occlusions and unpredictable dynamics in traffic. While adaptive techniques, such as adaptive Kalman filtering \cite{jiang2023novel}, can adjust parameters dynamically, intensive resources and handcrafted rules are necessary. Therefore, robust, adaptable algorithms that generalize across conditions, while supporting real-time processing, are still an open problem. \cite{fischer2023qdtrack} and \cite{wojke2017simple} replace traditional trackers with fully deep-learning-based alternatives. However, it increases computational complexity, potentially canceling the primary advantage of the model-based TBD framework, namely, real-time processing. As a result, such solutions risk becoming less practical than end-to-end joint detection and tracking approaches.}

{To address these limitations, we propose HybridTrack, a novel 3D multi-object tracking framework that combines the stability and interpretability of Kalman filters with the flexibility and generalization of deep learning. More specifically, HybridTrack uses a learnable Kalman filter to dynamically adjust stochastic parameters in real-time, eliminating manual tuning. 
With learnable components integrated in the motion model, Kalman gain, and noise covariances, the system can adapt autonomously to various scenarios without agent-specific designs and prior knowledge of the scene.
To do so, we design the deep-learning-based component with a limited number of parameters, leading to a lightweight framework. Additionally, the inherent recursive property of our learnable Kalman filter ensures data efficiency, while its learnable nature makes it data scalable. Furthermore, we introduce a dynamic scaling mechanism to enhance the stability of our model during noisy initialization. To summarize, our contributions are threefold: }
\begin{itemize}
    \item We propose a new 3D MOT framework, HybridTrack, based on a learnable Kalman filter by integrating novel deep learning modules. It allows us to combine the recursive property of Kalman filtering and the predictive capability of deep learning.
    \item A novel predictive motion model block and a Kalman Gain block are introduced. By doing so, HybridTrack eliminates the need for manual motion and stochastic parameter design.
    \item The experimental results demonstrate that the proposed method significantly outperforms the existing model-based tracking techniques on the KITTI dataset with $82.72\%$ HOTA. Furthermore, HybridTrack is able to achieve the fastest processing speed of 112 FPS, which unlocks the potential of real-time traffic applications.
\end{itemize}  
\section{Related Work}
\label{sec:format}
{\textit{Tracking-by-Detection (TBD}). TBD remains the most widely adopted paradigm in multi-object tracking (MOT) due to its modularity and computational efficiency. In TBD, an object detector \cite{shi2020pv, wu2023virtual, wu2022casa} generates detections, which are then processed by a separate tracking module for association and trajectory prediction. The decoupling of detection and tracking allows for flexibility, making TBD a practical choice for a wide range of applications. Traditional TBD methods often rely on statistical techniques such as Kalman filters \cite{welch1995introduction} or, more generally, Bayesian filters \cite{chen2003bayesian} for object association and trajectory prediction. These lightweight and interpretable approaches are ideal for real-time applications. However, their reliance on predefined motion models, such as constant velocity or constant acceleration assumptions \cite{wu20213d}, \cite{cao2023observation}, \cite{wang2023camo}, limits their ability to adapt to complex, dynamic scenarios. Extensions to non-linear motion models, such as Extended Kalman filters \cite{di2024monokalman}, have not demonstrated significant improvement in the aforementioned challenging conditions. More recent advancements, including Adaptive Kalman filters \cite{jiang2023novel} and probabilistic Bernoulli-based approaches \cite{article_pmbm} attempt to address these limitations. However intricate parameter tuning and significant domain knowledge are required, reducing these methods' scalability. Similarly, other works, including UCMCTrack \cite{yi2024ucmctrack} and PMTrack \cite{guo2024pmtrack}, aim to address challenges such as variable frame rates and complex sensor movements by incorporating advanced motion estimation techniques into motion particle filters. To further enhance robustness without abandoning the modular TBD paradigm, EMMS-MOT \cite{xu2024exploiting} and BiTrack \cite{huang2024bitrack} introduce a multi-modal fusion strategy that corrects 3D detections using 2D cues and jointly models motion across modalities. This approach improves resilience to occlusion and sensor noise, while retaining the flexibility of classic TBD designs. While real-time performance has always been a critical challenge for trackers generally, Fast-Poly \cite{li2024fast} introduces a polyhedral tracking framework that leverages rotated object alignment and dense local computation to significantly boost efficiency. It achieves promising speed-accuracy trade-offs on large-scale datasets with CPU-friendly design.}

{\textit{Data Association}. As a critical component of TBD, data association links detections across frames to maintain vehicle identities. Traditional methods like AB3DMOT \cite{weng20203d} rely on spatial metrics, Intersection over Union (IoU) and Euclidean distance, respectively. While simple and efficient, these metrics struggle in complex scenarios, leading to identity switches and trajectory fragmentation. To address these issues, advanced approaches incorporate Confidence-guided association \cite{wu20213d}, \cite{he20243d} and appearance-based features \cite{wang2023camo}. Recent works also integrate multi-stage association strategies, improving robustness in conditions such as occlusion and low-confidence detections \cite{wang2022deepfusionmot}. However, the dependency on empirically defined metrics limits the adaptability to varying environments. Deep learning has further revolutionized TBD by enabling data-driven feature extraction and association. DeepSORT \cite{wojke2017simple} uses neural networks for robust appearance modeling. TripleTrack \cite{marinello2022triplettrack} and AppTracker++ \cite{zhou2024apptracker+} fuse detection and tracking cues at the feature level to enhance robustness. DINO-MOT \cite{lee2024dino} enhances pedestrian tracking in the TBD framework by integrating a visual foundation model (DINOv2) for pedestrian re-identification. It introduces a visual memory mechanism to reduce ID switches and leverages appearance features from the camera modality. Another promising direction involves the use of graph theory for association. PolarMOT \cite{kim2022polarmot} employs spatial and temporal graph-based strategies leveraging 3D geometric cues, while \cite{wang2023towards} applies graph theory as a post-processing step for trajectory rectification. However, deep learning-based approaches often face challenges related to high data dependency and limited interpretability, making them less generally practical. One can see that most methods leveraging deep learning for multi-object tracking utilize it in either an end-to-end pipeline, where both detection and tracking are integrated, as seen in JDT \cite{tokmakov2021learning}, \cite{tokmakov2022object},\cite{wu2021tracklet}, \cite{10777493} or within a learnable data association step \cite{wojke2017simple}, \cite{marinello2022triplettrack}, \cite{zhou2024apptracker+}.} In contrast, HybridTrack bridges the gap between these paradigms by embedding the predictive capabilities of deep learning into a traditional Kalman filter framework followed by a one-step greedy-algorithm-based data association. This hybrid approach enables a lightweight and interpretable network that maintains robustness and data efficiency. By combining the adaptability of deep learning with a Kalman filter structure, HybridTrack effectively addresses the limitations of purely traditional or fully deep learning-based methods, providing a scalable and resource-efficient solution for MOT tasks.
\section{HybridTrack}
\begin{figure*}[h]
\centering
\includegraphics[width=\textwidth]{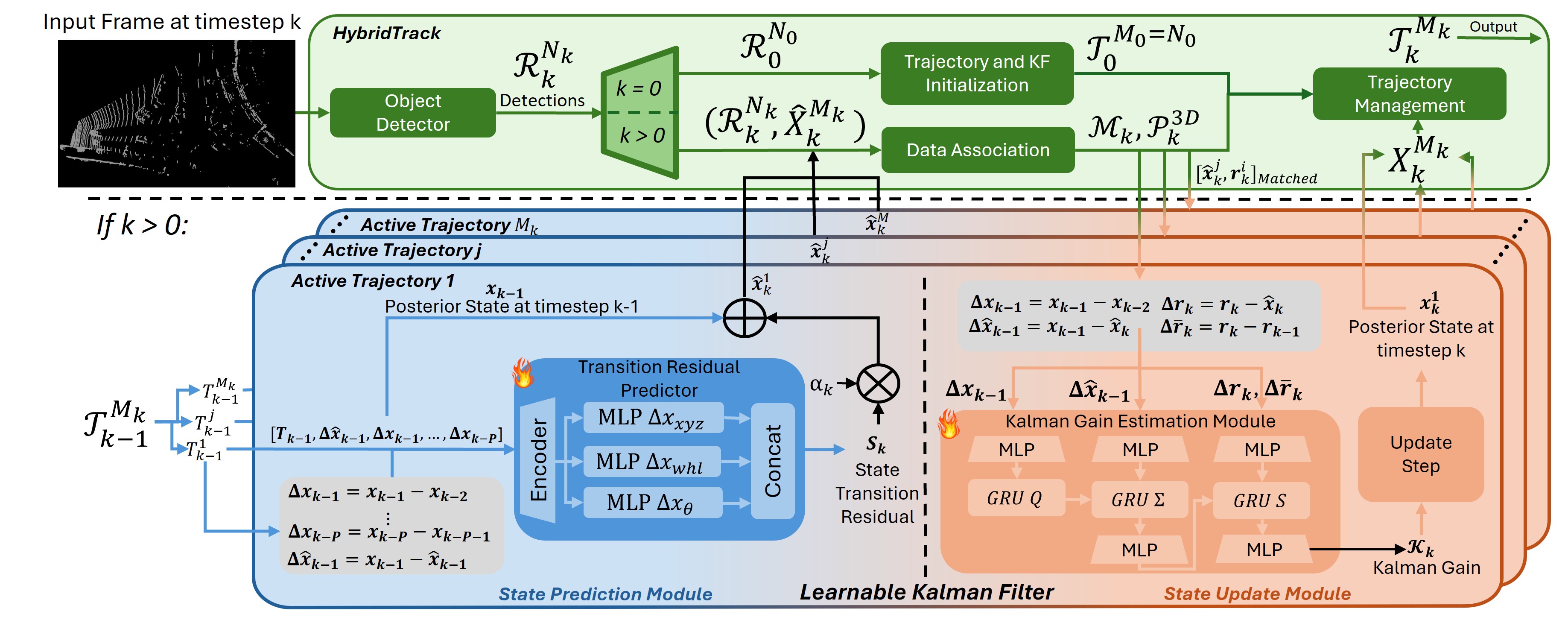}
    \caption{The proposed HybridTrack. During inference, HybridTrack begins with 3D object detection, localizing vehicles using sensor data. The resulting detections $\mathcal{R}^{N_k}_k$ are sent to the State Prediction Module (SPM), where new detections $\mathbf{r}^i_k$ initialize trajectories $\mathcal{T}_{k}^{M_k}$. For each trajectory $T_{k}^j$, the TRP predicts the next state $\hat{\mathbf{x}}^j_{k} = \alpha_k S^j_{k} + \mathbf{x}^j_{k-1}$ forming the set of predicted states $\hat{X}_{k}^{M_k}$. In the Data Association module, $\hat{X}_{k}^{M_k}$ is matched with current $\mathcal{R}^{N_k}_k$. The State Update Module (SUM) refines the matched pairs, updating the vehicle states ${X}_{k}^{M_k}$. Finally, the Trajectory Management module maintains and updates trajectories.}
    \label{fig:architecture}
\vspace{-0.5cm}
    
\end{figure*}
\subsection{Problem Formulation and Architecture Overview}
Given a sequence of video/LiDAR point cloud frames from a real-world traffic scene, our goal is to accurately and efficiently identify and track multiple vehicles in 3D space. To do so, our tracking-by-detection framework consists of five key modules, shown in Fig. \ref{fig:architecture}:
1) Object Detection: A 3D detection model extracts 3D bounding boxes from sensor data (e.g., video or LiDAR) at each timestep, identifying and localizing vehicles in the 3D world. 2) Trajectory Initialization and State Prediction: Newly detected vehicles initialize a trajectory, and the State Prediction Module (SPM) predicts the next state of each tracked trajectory. 3) Data Association: Detected vehicles from (1) are matched with predicted states from (2). 4) State Update: Vehicle trajectories are refined and updated based on the associations. 5) Trajectory Management: Handles and maintains trajectories for consistent tracking. 
In the following sections, we elaborate on each component.
\subsection{3D Object Detection}
 The fundamental concept of the tracking-by-detection approach is to track objects i.e., vehicles in our case, using detections at each timestep $k$, extracted by a 3D object detector. Formally, the detector $\mathcal{F}$ outputs a set $\mathcal{R}^{N_k}_k = \mathcal{F}(I_k)$, where $I_k$ is the input data frame at timestep $k$, and $\mathcal{R}^{N_k}_k = \{\mathbf{r}^i_k\}_{i=0}^{N_k} \in \mathbb{R}^{N_k \times F}$ represents the detected vehicles. Here, $N_k$ is the number of detections at timestep $k$, $F$ is the number of features, and each $\mathbf{r}^i_k \in \mathbb{R}^F$ is the $i$-th vehicle. In other words, each measurement $\mathbf{r}^i _k$ corresponds to a 3D bounding box. In this paper, measurements are defined as $\mathbf{r}^i_k = [x, y, z, w, l, h, \theta] \in \mathbb{R}^7$, where $x, y, z$ are the 3D centroid coordinates, $w, l, h$ are width, length, and height, and $\theta$ is the heading angle.
 
\subsection{Trajectory Initialization}
In HybridTrack, trajectories are classified as active or dead. Active trajectories represent currently tracked vehicles, while dead ones correspond to vehicles undetected for $\mu_{\text{max}}$ consecutive timesteps. At $k=0$, each detection $\mathbf{r}^i_0 \in \mathcal{R}^N_0$ initializes an active trajectory $T_{0}^j$, forming a one-to-one mapping. The set of active trajectories is defined as $\mathcal{T}_{k}^{M_k} = \{T_{k}^j\}_{j=0}^{{M_k}}$, where $M_k$ is the number of tracked vehicles at timestep $k$. Initially, $D_0 = \emptyset$, and the set of dead trajectories $D_k$ is updated when new ones are created. At $k > 0$, each $T_k^j$ includes the current detection $\mathbf{r}_k^i$, predicted state $\hat{\mathbf{x}}_k^j$, and past posterior states $\mathbf{x}^j_{k-1} \ldots,  \mathbf{x}^j_{k_0}$, forming $T_{k}^j = [\mathbf{r}_k^i,\hat{\mathbf{x}}_k^j,\mathbf{x}^j_{k}, \mathbf{x}^j_{k-1}\ldots,  \mathbf{x}^j_{k_0}] \in \mathbb{R}^{L \times F}$, where $L = k - k_0 + 2$ and $k_0$ is the trajectory’s start time. At $k = 0$, indices $i$ and $j$ are identical. When initializing a trajectory during inference, prior states $\mathbf{x}^j_{k_0-1} \ldots,  \mathbf{x}^j_{0}$ are set to $\mathbf{x}_{k_0}^j - \epsilon$ to avoid duplicates. A learnable Kalman filter (LKF) is also initialized per trajectory, with its prior and posterior estimates.

\subsection{State Prediction Module} 
At each timestep $k > 0$, the State Prediction Modules (SPMs), integrated in each LKF, predict the next state of the associated tracked trajectories. Rather than computing the prior state directly, we model the Kalman filter’s prediction step as a deep-learning problem.
\subsubsection{Mathematical Formulation}
Let $\mathbf{x}_{k}^j \in \mathbb{R}^F$ denotes the posterior state vector of a j-th tracked vehicle at timestep $k$, which includes the 3D location, 3D dimensions, and orientation of the vehicle. We aim to learn a function $f_\theta(\mathbf{x}^j_{k-1}): \mathbb{R}^F \to \mathbb{R}^F$, parameterized by $\theta$, that maps the previous posterior state $\mathbf{x}_{k-1}^j$ to the predicted prior state $\hat{\mathbf{x}}^j_k$. To mitigate the error probability and introduce more stability to the system, instead of directly predicting the prior state $\hat{\mathbf{x}}^j_k$, we aim to predict the state transition residual $S^j_{k}$, which captures the state deviation. This enables the model to focus on learning motion dynamics rather than full state estimation. The network is trained to minimize the prediction error relative to the true posterior state at timestep $k$., formulating the problem as supervised learning with a loss function. 

\subsubsection{Transition Residual Predictor (TRP)}
We parametrize the function \( f_\theta(\mathbf{x}_{k-1}^j)\) using an encoder-decoder architecture. It processes the historical trajectory information of a tracked vehicle to predict its transition residual, as illustrated in Fig. \ref{fig:architecture}. The SPM is lightweight and parameter-efficient, as we are leveraging the recursive architecture of the deep learnable Kalman filter to effectively correct small errors and maintain accurate predictions with minimal computational overhead. Each SPM is applied independently to each active trajectory, enabling the TRP module to adaptively generate transition residuals tailored to the specific motion patterns of individual vehicles. To effectively capture the temporal dynamics of vehicle motion, the TRP leverages input that includes the current trajectory $T_{k-1}^j$ and, for any integer $0 \leq P \leq k-k_0-1$, the sequence of posterior state differences $[\Delta \mathbf{x}_{k-1}^j, \ldots, \Delta \mathbf{x}_{k-P}^j] \in \mathbb{R}^{P \times F}$ each defined as $\Delta \mathbf{x}_{k-i}^j = \mathbf{x}_{k-i}^j - \mathbf{x}_{k - i - 1}^j, \quad \text{for } 1 \leq i \leq P$. Additionally, the Kalman gain update term \(\Delta \hat{\mathbf{x}}_{k-1}^j\) is included. This term incorporates information on the correction performed by the Kalman filter during the previous timestep. 
The TRP encodes its input variables, $[T^j_{k-1}, \Delta \hat{\mathbf{x}}^j_{k-1}, \Delta \mathbf{x}^j_{k-1}, \ldots, \Delta \mathbf{x}^j_{k-P}],$ using an encoder to extract features. Subsequently, three parallel Multi-Layer Perceptrons (MLPs) compute the residuals for specific vehicle states: (1) Spatial information $\Delta \mathbf{x}^j_{\text{xyz}}  = [\Delta x, \Delta y, \Delta z]$ which represents the residuals in 3D coordinates of the vehicle, (2) Geometrical properties $\Delta \mathbf{x}^j_{\text{whl}} = [\Delta w, \Delta l, \Delta h]$ representing the residuals in dimension of the 3D bounding box, and (3) the residual in vehicle orientation $\Delta \theta$. The final state transition residual $S^j_{k}$ is computed as: $S_k^j = \begin{bmatrix} \Delta x & \Delta y & \Delta z & \Delta w & \Delta l & \Delta h & \Delta \theta \end{bmatrix}$. 
This setup helps capture individual vehicle behavior and maintain stable residual predictions. 
\subsubsection{Dynamic Scaling Factor}
One limitation of the LKF is its need for several timesteps to calibrate noise parameters, which complicates prediction stabilization when initial detections are noisy due to sensor errors, occlusions, or environmental factors. This may cause prediction overshooting, with updates that deviate excessively from the true trajectory, leading to missed associations.
To mitigate this, we introduce an adaptive scaling factor $\alpha$ that dampens early prediction updates. By reducing the impact of abrupt changes, $\alpha$ limits unstable corrections and allows the Kalman filter to gradually calibrate noise parameters and improve robustness and accuracy. It is defined by the time elapsed since initialization $\tau_0$:
\begin{equation}
    \alpha_k = \frac{1}{T_{\text{max}}} \times (k - \tau_0), \text{ if } (k - \tau_0) < T_{\text{max}},
\end{equation}
otherwise, $\alpha_k=1$. This linear schedule smooths initial predictions and mitigates overshooting.  As $\alpha$ approaches 1, the Kalman filter transitions to its standard behavior, relying on calibrated noise parameters for accurate trajectory estimation. 

On the other hand, missed detections are an additional challenge for the SPM, particularly in dense traffic scenarios. More specifically, it forces the LKF to rely solely on the process model during prediction, leading to compounding errors and potential divergence.  Prior work \cite{wu20213d, he20243d} addresses this by scaling the cost function with confidence scores, expanding the search radius during data association. However, this increases the risk of false positives and ID switches in crowded scenes. Moreover, such scaling is limited to linear cost functions like Euclidean distance. Instead of modifying the cost function, we propose mitigating the impact of missed detections by directly adjusting the predicted state via a refined scaling factor $\alpha$. The factor is dynamically adjusted as follows:
\begin{equation}
\alpha_k = \alpha_k \times \left(1 - \frac{\kappa}{\mu_{\max}}\right) \cdot (a_{\max} - a_{\min}), \text{ if } \kappa < \mu_{\max}, \\
\end{equation}
Otherwise, $\alpha_k = a_{\min}$. Here, $\kappa$ is the number of consecutive missed detections, $\mu_{\max}$ is the threshold before a trajectory is marked dead, and $a_{\max}, a_{\min}$ are empirically chosen bounds.
By reducing $\alpha$ dynamically, our method stabilizes tracking in both early stages and in cases of missed detections, minimizing the impact of noisy predictions while maintaining robust data association.
\subsubsection{Prior State Prediction}
Using the predicted transition residual $S^j_k$ and the scaling factor $\alpha_k$, the Kalman filter prior state can be defined as follows:
\begin{equation}
\begin{aligned}
    \hat{\mathbf{x}}^j_{k} &= \alpha_k f_\theta\left(T^j_{k-1}, \Delta \mathbf{x}^j_{k-1}, \Delta \mathbf{x}^j_{k-2}, \ldots, \Delta \mathbf{x}^j_{k-P}, \Delta \hat{\mathbf{x}}^j_{k-1}\right) + \mathbf{x}^j_{k-1}, \\
    &= \alpha_k S^j_{k} + \mathbf{x}^j_{k-1}.
\end{aligned}
\label{eq:KF_state_pred}
\end{equation}
with $f_\theta\left(\cdot\right)=TRP(\cdot)$. Therefore, for each active trajectory $T_{k-1}^j$, a predicted prior state is inferred. We define the set of predicted prior states as follows: ${\hat{X}}_{k}^{M_k} = \{\hat{\mathbf{x}}^j_{k}\}_{j=0}^{{M_k}} \in \mathbb{R}^{{M_k} \times F}$.
\subsection{Data Association}\label{sec:data_ass}
Data association links current detections $\mathcal{R}^{N_k}_k$ to active tracked trajectories $\mathcal{T}_{k}^{M_k}$, ensuring consistent vehicle trajectories over time. For this step, we propose a one-stage matching process. 
First, a cost map is computed, defined as $C_{3D}({\hat{X}}_{k}^{M_k}, \mathcal{R}^{N_k}_{k}) = 1 - CIoU({\hat{X}}_{k}^{M_{k}}, \mathcal{R}^{N_k}_{k})$ where $CIoU$ is the Complete Intersection over Union (CIoU) \cite{zheng2021enhancing}. The cost map $C_{3D}$ is a $M_{k} \times N_k$ matrix, with each element  $C_{3D}[j,i]$ corresponding to the cost of associating the $j-th$ predicted vehicle with the $i-th$ detected vehicle. 
By using $C_{3D}$, we apply a simple yet effective greedy algorithm with a fixed threshold $\tau_{3D}$ to associate the newly detected vehicles $r^i_{k} \in \mathcal{R}^{N_k}_{k}$ with the active trajectories $T_{k-1}^j$ and their corresponding prediction states $\hat{\mathbf{x}}^j_{k}$. The greedy algorithm assigns detections to trajectories by iteratively selecting the lowest-cost $C_{3D}[j,i]$ for the pair ($\hat{\mathbf{x}}^j_k,\mathbf{r}^i_k$) at each step, $\tau_{3D}$ serves as a criterion for deciding whether detection and a trajectory prediction are close enough to be associated. This association step returns the matched indices $(j,i)$ and the set of matched pairs $\mathcal{P}_k^{3D} \in \mathbb{R}^{J_k \times 2 \times F}$ where $J_k$ is the number of matched pairs at timestep k with $J_k \leq M_{k}$ and one element is represented as $(\hat{\mathbf{x}}^j_{k}, \mathbf{r}^{i}_{k})$. 
\subsection{State Update Module}
For every matched pair $(\hat{\mathbf{x}}^j_{k}, \mathbf{r}^{i}_{k})$ in $\mathcal{P}_k^{3D}$, we update the previous posterior state $\mathbf{x}^j_{k-1} $ of the corresponding trajectory $T_{k-1}^j$ in the State Update Module (SUM). Our updated step is inspired by KalmanNet \cite{revach2022kalmannet}. That is, instead of relying on a traditional Kalman filter, \textit{HybridTrack} replaces the Kalman gain $\mathcal{K}^j_k \in \mathbb{R}^{F \times F}$ and the covariances $\mathbf{v_k}$ and $\mathbf{w_k}$ with a Recurrent Neural Network (RNN) \cite{elman1990finding} based architecture. This architecture is described in Fig. \ref{fig:architecture} dubbed Kalman Gain Estimation Module. Once the Kalman gain  $\mathcal{K}^j_k$ is inferred, we can compute the posterior state as follows:
\begin{equation}
    \text{Updated State Estimate} : \mathbf{x}^j_{k} = \hat{\mathbf{x}}^j_{k} + \mathcal{K}^j_k (\mathbf{r}^i_{k} - H \hat{\mathbf{x}}^j_{k}),
    \label{eq:updated_state_tracker}
\end{equation} 
\noindent where $\mathbf{\hat{x}}^j_{k}$ is the predicted state estimate, $\mathbf{r}^i_{k}$ denotes the matched detection/measurement with the associated indexes $(j,i)$. $H \in \mathbb{R}^{F \times F}$ is the observation model mapping state $\hat{\mathbf{x}}^j_{k}$ to measurement space with $H: \mathbb{R}^F \to \mathbb{R}^F, \quad \mathbf{r}_k^i \approx H \mathbf{x}_k^j$. Here, $H$ is defined as an identity matrix. On the other hand, the Kalman gain $\mathcal{K}_k^j$ effectively maps $(\mathbf{r}^i_{k} - H \mathbf{\hat{x}}^j_{k})$ from the measurement space \( \mathbb{R}^F \) back into the state space \( \mathbb{R}^F \). 
 For all updated posterior states $\mathbf{x}^j_{k}$ of the active trajectories $\mathcal{T}_{k-1}^{M_k}$, we aggregate them into a set of updated detections, ${{X}}_{k}^{M_k} = \{{\mathbf{x}}^j_{k}\}_{j=0}^{M_k} \in \mathbb{R}^{{M_k} \times F}$, which is then passed to the trajectory management module.
The main advantage of this approach is that it eliminates the need to manually model state and measurement noise parameters $\mathbf{v_k}$ and $\mathbf{w_k}$, which are often nonadaptive for diverse and dynamic scenes. Instead, our method employs dynamically changing covariances rather than static, Gaussian-distributed noise parameters.
\subsection{Trajectory Management} 
Active trajectories are updated with the matched updated posterior states ${{X}}_{k}^{M_k}$. For unmatched detections $\mathcal{R}^{N_k}_k$, new trajectories are initialized to account for newly appearing vehicles and added to the set of active trajectories. If a predicted state $\hat{\mathbf{x}}_k^j \in \hat{X}_{k}^{M_{k}}$ of an active trajectory $T_{k-1}^j \in \mathcal{T}^{M_{k}}_{k-1}$ fails to be associated with a detection $\mathbf{r}_k^i \in \mathcal{R}^{N_k}_k$, the trajectory $T_{k-1}^j$ is marked as non-updated. Moreover, the trajectory is classified as dead and added to $D_{k}$ after $\mu_{max}$ consecutive timesteps without updates. Additionally, if a new trajectory is wrongly created due to a detection artifact, it will be designated dead if it remains non-updated for $\sigma$ timesteps.
\subsection{Optimization}\label{sec:optimization}
The proposed learnable Kalman filter is trained independently from the tracking pipeline using loss functions that enforce spatial and temporal consistency. That is, the SPM and SUM are trained together in an end-to-end manner. 
\begin{table*}[h]
\caption{{Comparison of existing methods on the KITTI test set. The best is marked in \textbf{bold}, and the second-best in \underline{underline}. The results are reported in \%.}}
\label{tab:comparison_of_methods}
\resizebox{\linewidth}{!}{%

    \begin{tabular}{lcc|ccc|ccc|c}
    \toprule
     & & & \multicolumn{3}{c|}{\textbf{Detection}} & \multicolumn{3}{c|}{\textbf{Association}} &  \\
    \textbf{Method}&\textbf{Modality} & \textbf{HOTA} $\uparrow$ & \textbf{DetA} $\uparrow$  & \textbf{DetRe}  $\uparrow$ & \textbf{DetPr} $\uparrow$ & \textbf{AssA} $\uparrow$ & \textbf{AssRe} $\uparrow$ & \textbf{AssPr} $\uparrow$ & \textbf{LoCA}  $\uparrow$ \\
    \midrule
PermaTrack \cite{tokmakov2021learning} \textit{(ICCV'21)} & 2D &78.03 &78.29 &81.71 &{86.54} &78.41 &81.14 &89.49 &87.10 \\
PC-TCNN \cite{wu2021tracklet} \textit{(IJCAI'21) } & 3D &80.90 &78.4 & {84.22} &84.58 &84.13 &87.46 &{90.47} &87.48 \\
TripletTrack \cite{marinello2022triplettrack} {(CVPR'22)} & 2D & 73.58  & 73.18   & 76.18  & {86.81}  & 74.66  & 77.31  & 89.55  & 87.37  \\
RAM \cite{tokmakov2022object} \textit{(ICML'22)} & 3D &79.53 &{78.79} &82.54 &86.33 &80.94 &84.21 &88.77 &87.15 \\
DF-MOT \cite{wang2022deepfusionmot} \textit{(RAL'22)} & 2D+3D & 75.46  & 71.54  & 75.34  & 85.25  & 80.05  & 82.63   & 89.77  & 86.07  \\
RethinkingMOT \cite{wang2023towards}   \textit{(ICRA'23)} & 3D &80.39 &77.88 & \underline{84.23} &83.57 &83.64 &{87.6} &88.90 &87.07 \\
OC-SORT \cite{cao2023observation}\textit{(CVPR'23)} & 2D &76.54 &77.25 &80.64 &86.34 &76.39 &80.33 &87.17 &87.01 \\
FNC2 \cite{jiang2023novel} \textit{(TIV'23)}& 2D+3D &73.19 &73.27 &80.98 &81.67 &73.77 &77.05 &89.84 &87.31 \\
AppTracker \cite{zhou2024apptracker+} \textit{(IJCV'24)} & 2D &75.19 &75.55 &78.77 &86.04 &75.36 &78.34 &88.24 &86.59 \\
PNAS-MOT \cite{peng2024pnas} \textit{(RAL'24)} & 2D+3D &67.32 &77.69 &81.58 &85.81 &58.99 &64.70 &80.74 &86.94 \\
UCMCTrack \cite{yi2024ucmctrack} \textit{(AAAI'24)}& 2D & 77.10  & -& -& -&77.20  & -& -& -  \\
UG3DMOT \cite{he20243d} \textit{(SP'24)} & 3D & 78.60  & 76.01  & 80.77  & 85.44  & 82.28  & 85.36  & { 91.37}   & {87.84}  \\
MMF-JDT \cite{10777493} \textit{(RAL'24)} & 2D+3D & 79.52  & 75.83  & 82.31  & 83.69  & 84.01  & 87.16  & 90.70   & 87.65  \\
PMTrack \cite{guo2024pmtrack} \textit{(ACCV'24)}  & 3D & {81.36}  & {78.90}  & 82.98  & {86.76}  & {84.49}  & {87.73}  & {90.18}  & {88.02}   \\
MCtrack \cite{wang2024mctrack} \textit{(ArXiv'24)}  & 3D & \textbf{82.75} & \underline{79.40} & 82.92 &	\textbf{87.45} & \underline{86.85}  &	\underline{89.84} &	\underline{91.32} &	\underline{88.07}   \\
BiTrack \cite{huang2024bitrack} \textit{(ArXiv'24)}  & 2D+3D & {82.70} &	\textbf{80.04} &		\textbf{84.53} &	87.19 &	86.17& 89.11 &	\textbf{92.16 }&	\textbf{88.68} \\
\hline
HybridTrack \textit{(Ours)} & 3D & \underline{82.72} &	79.28 &	82.84 &	\underline{87.39} & \textbf{86.92} &	\textbf{89.97} & {	91.24} & {88.04}  \\
\bottomrule
	\end{tabular}}
    \label{tab:main_results}
\vspace{-0.5cm}

\end{table*}
Specifically, HybridTrack is trained by using fixed-length (20 frames) ground-truth tracklets. During training phase, we use ground-truth pose sequences. No data association or occlusion handling is required. The prior and posterior state predictions are optimized by L1 loss. In contrast, no significant impact on performance is observed when augmenting model sequences input with noise. Therefore, we mainly focus on learning accurate motion dynamics by using the original clean data, while the Kalman filter handles uncertainty during inference.

The primary loss, Mean Absolute Error (MAE), minimizes errors in both prior and posterior predicted states, denoted as $\mathcal{L}_{\text{states}}$. To ensure temporal smoothness, a temporal consistency loss $\mathcal{L}_{\text{temporal}}$ penalizes large deviations between consecutive states. A direction consistency loss $\mathcal{L}_{\text{direction}}$ further stabilizes movement by normalizing speed vectors to avoid abrupt directional changes. Let \(\bar{\mathbf{x}}_{i,k}\), \(\hat{\mathbf{x}}_{i,k}\) and \(\mathbf{x}_{i,k}\) denote the ground truth, predicted prior and posterior sequences for the single $i$-th vehicle, respectively, with $k$ indexing the temporal dimension and $i$ the data sample index. 
The losses are formulated as follows:
  \begin{equation}
   \mathcal{L}_{\text{states}} =\frac{1}{N_D} \sum_{i=1}^{N_D} \frac{1}{T} \sum_{k=1}^{T} \left(\|\bar{\mathbf{x}}_{i,k} - {\mathbf{x}}_{i,k}\|_1 +  \|\bar{\mathbf{x}}_{i,k} - \hat{\mathbf{x}}_{i,k}\|_1\right)
   \end{equation}
    \begin{equation}
        \mathcal{L}_{\text{temporal}} = \frac{1}{N_D} \sum_{i=1}^{N_D}\frac{1}{T-1} \sum_{k=1}^{T-1}  \|\mathbf{p}_{i,k+1} - \mathbf{p}_{i,k}\|_1
    \end{equation}
   \begin{equation}
   \mathcal{L}_{\text{direction}} = \frac{1}{N_D} \sum_{i=1}^{N_D}\frac{1}{T-1} \sum_{k=1}^{T-1} \|\mathbf{d}_{i,k+1} - \mathbf{d}_{i,k}\|_2
    \end{equation}
   where  $N_D$ is the dataset size, $\mathbf{p}_{i,k} = [x,y,z]$ is the centroid of the vehicle, $\mathbf{d}_{i,k} = \frac{\mathbf{v}_k}{\|\mathbf{v}_k\| + \epsilon}$ is the normalized direction vector at time \(k\), and \(\mathbf{v}_k\) is the speed vector. $\epsilon>0$ is a small constant added to the denominator to avoid division by zero.

\textbf{Joint loss optimization.} The total loss is a weighted sum of these individual losses:
\begin{equation}
    \mathcal{L}_{\text{total}} = \lambda_{\text{states}} \mathcal{L}_{\text{states}} + \lambda_{\text{temp}} \mathcal{L}_{\text{temporal}} + \lambda_{\text{dir}} \mathcal{L}_{\text{direction}}
\end{equation}
Where we set $\lambda_\theta = 1$, for all $\theta \in \{states, temp, dir\}$. This loss function ensures the model maintains consistency in the predicted sequences' spatial and temporal dimensions.

\section{Experiments}
\textbf{\textit{Dataset and Implementations}}. The KITTI dataset \cite{geiger2012we} is a widely used dataset for monocular object pose estimation and tracking. The dataset comprises 50 videos, divided into 21 for training and 29 for testing, following the recommended splitting on the official website. Following the standard practice in the KITTI benchmark, the training dataset is further subdivided into 10 sequences for training and 11 sequences for validation. To train  \textit{HybridTrack}, we extract the pose predictions from an existing 3D bounding box estimator as $\mathbf{r}_k$. We use VirConv \cite{wu2023virtual} as a 3D object detector to predict 3D bounding boxes. The network is optimized by Adam Optimizer \cite{kingma2017adam} with a learning rate of 0.001 and a weight decay of 0.00001. {We take a batch size of 128 on one Nvidia Titan X (12G). The iteration number for the training process is set to 1,500. During inference, $T_{\text{max}} = 8$, $a_{\text{min}} = 0.1$, $a_{\text{max}} = 0.9$, $\mu_{\text{max}} = 22$, and $\sigma = 5$. The CIoU threshold for data association is set to 1.20.}

\textbf{\textit{Evaluation Metrics}}. We primarily focus on the Higher Order Tracking Accuracy (HOTA) metric \cite{luiten2021hota}, providing a balanced measure of detection, association, and localization accuracy in multi-object tracking. It is defined as $\text{HOTA} = \sqrt{\text{DetA} \cdot \text{AssA}}$.
DetA is the Detection Accuracy \cite{luiten2021hota} emphasizing the accuracy of object detection, while AssA is the Association Accuracy \cite{luiten2021hota} explicitly measuring the association's effectiveness in maintaining track consistency. We also furnish the Localization Accuracy (LoCA) which describes how accurately the objects' spatial positions are estimated. Additionally, we include the Classification of Events, Activities, and Relationships  Multi-Object Tracking (CLEARMOT) \cite{bernardin2008evaluating} metrics, such as MOTA \cite{bernardin2008evaluating} for Multi-Object Tracking Accuracy
and IDF1 \cite{ristani2016performance} for identity consistency, to complement the HOTA evaluation. To further evaluate our tracker, we also analyze MT (Mostly Tracked), PT (Partially Tracked), and ML (Mostly Lost) rates, which highlight the tracker's performance over time. Specifically, MT is the percentage of vehicle tracks that are successfully tracked for more than 80\% of their trajectory length, whereas ML is less than 20\%.
\begin{table}[h] 
\caption{Further Comparison based on CLEAR metric. $^1$ Uses CasA \cite{wu2022casa} detector. $^2$ Uses VirConv \cite{wu2023virtual} detector. \label{tab:clear_metrics}. The results are reported in \%.}
\resizebox{\columnwidth}{!}{%
\centering
\begin{tabular}{lccccc}
\toprule
\textbf{Methods}  & \textbf{MOTA} $\uparrow$ & \textbf{MT} $\uparrow$ & \textbf{PT} $\downarrow$ & \textbf{ML} $\downarrow$  & \textbf{IDFN} $\downarrow$\\
\midrule
$\text{UGD3MOT}^1$ \cite{he20243d}& 87.98  & {79.08} &{15.54 }&{\textbf{5.38 }}&1111\\
$\text{HybridTrack}^1$ & {90.16 }  & {84.61} &{9.23} &{6.15 }&{1350}\\
$\text{HybridTrack}^2$ & \textbf{91.05}  & \textbf{85.39} &\textbf{6.31} &{8.31 }&\textbf{534}\\
\bottomrule
\end{tabular}}
\vspace{-0.1cm}
\end{table}

\begin{figure*}[h!]
    \centering
    
   \begin{minipage}[b]{0.33\linewidth}
        \centering
        \includegraphics[width=\linewidth,trim={0cm 3.8cm 0cm 0.5cm},clip]{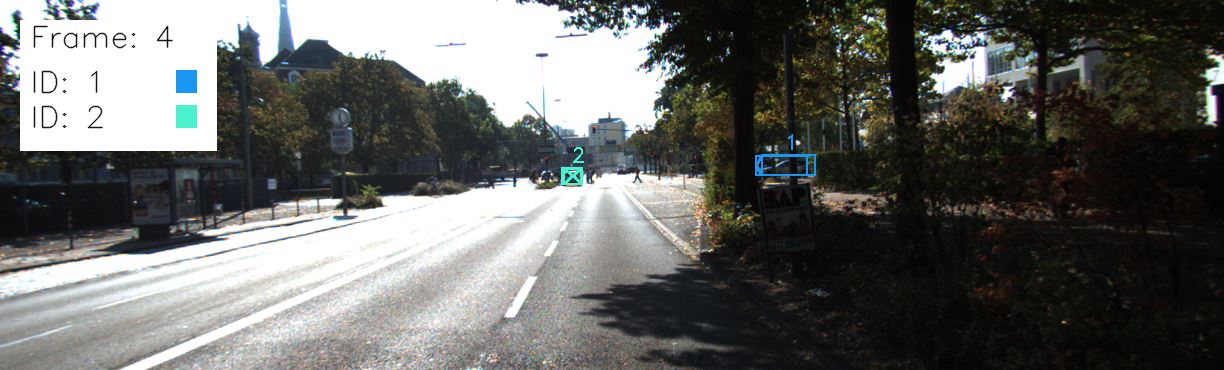}
    \end{minipage}%
    \begin{minipage}[b]{0.33\linewidth}
        \centering
        \includegraphics[width=\linewidth,trim={0cm 3.8cm 0cm 0.5cm},clip]{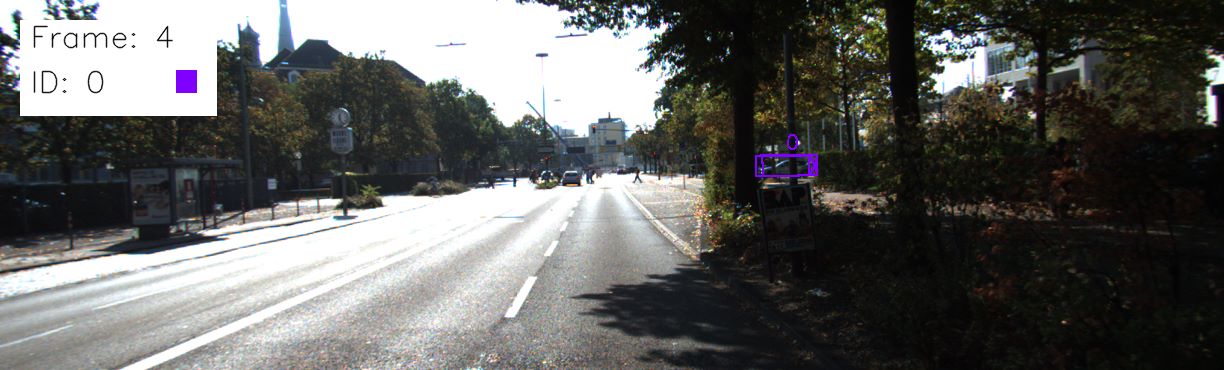}
    \end{minipage}%
    \begin{minipage}[b]{0.33\linewidth}
        \centering
        \includegraphics[width=\linewidth,trim={0cm 3.8cm 0cm 0.5cm},clip]{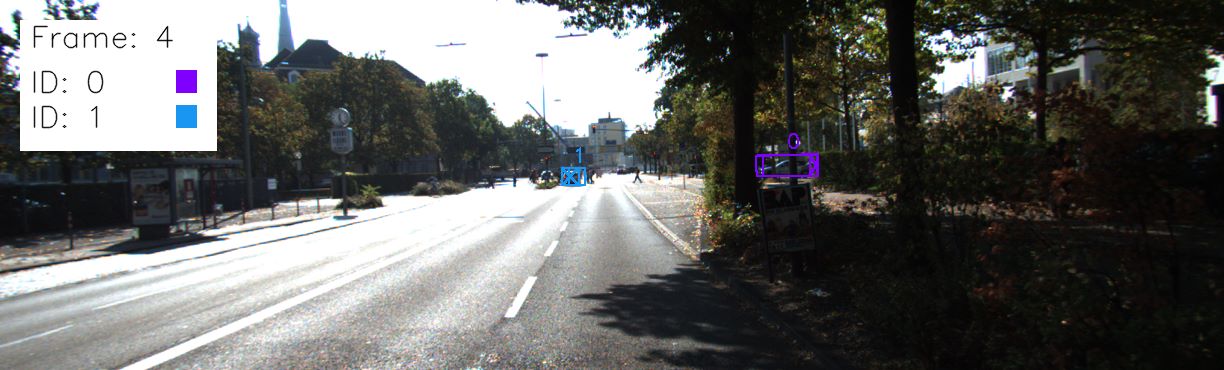}
    \end{minipage}\\
    \begin{minipage}[b]{0.33\linewidth}
        \centering
        \includegraphics[width=\linewidth,trim={0cm 3.8cm 0cm 0.5cm},clip]{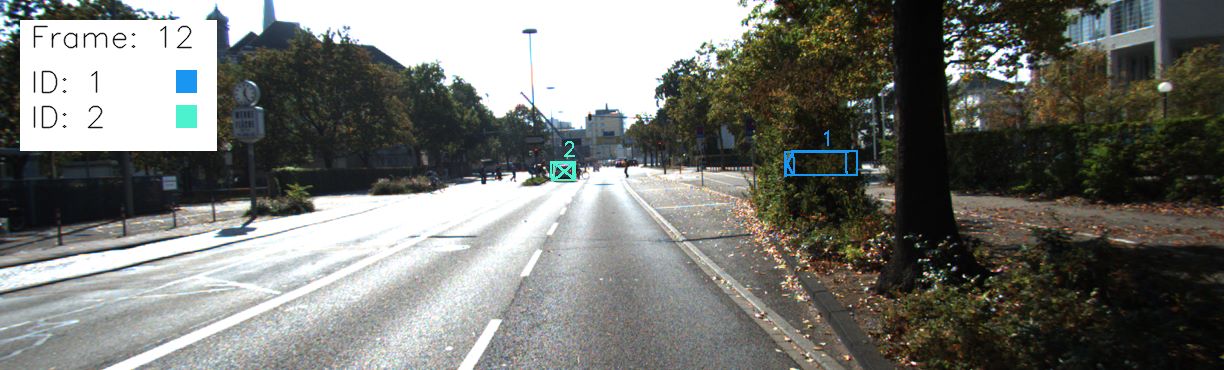}
    \end{minipage}%
    \begin{minipage}[b]{0.33\linewidth}
        \centering
        \includegraphics[width=\linewidth,trim={0cm 3.8cm 0cm 0.5cm},clip]{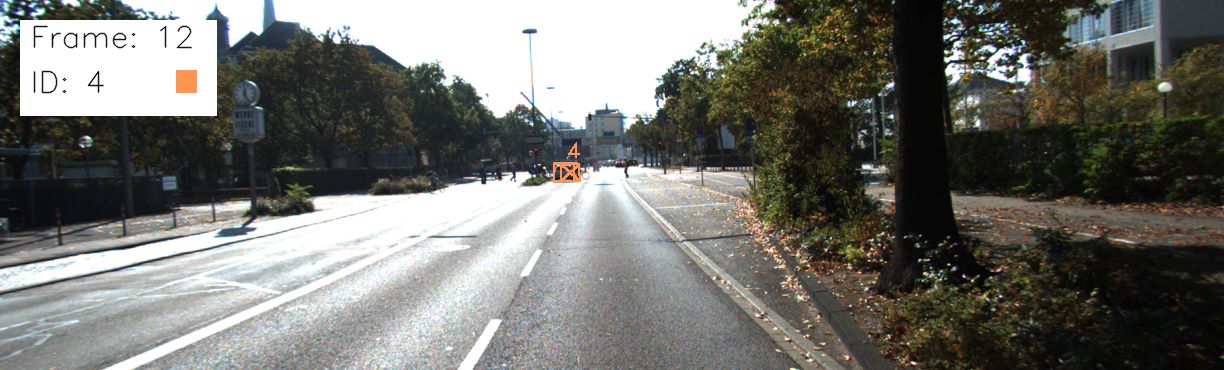}
    \end{minipage}%
     \begin{minipage}[b]{0.33\linewidth}
        \centering
        \includegraphics[width=\linewidth,trim={0cm 3.8cm 0cm 0.5cm},clip]{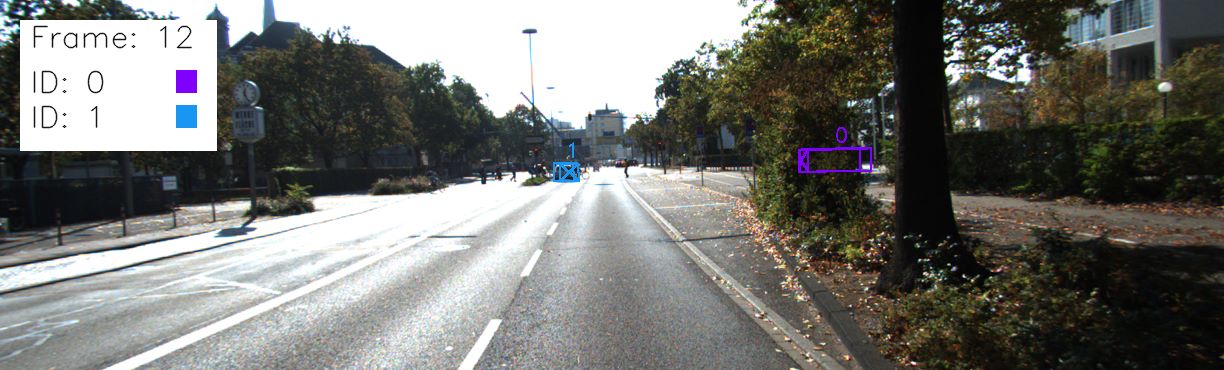}
    \end{minipage}\\[0.01cm]
    \begin{minipage}[b]{0.33\linewidth}
        \centering
        \includegraphics[width=\linewidth,trim={0cm 3.8cm 0cm 0.5cm},clip]{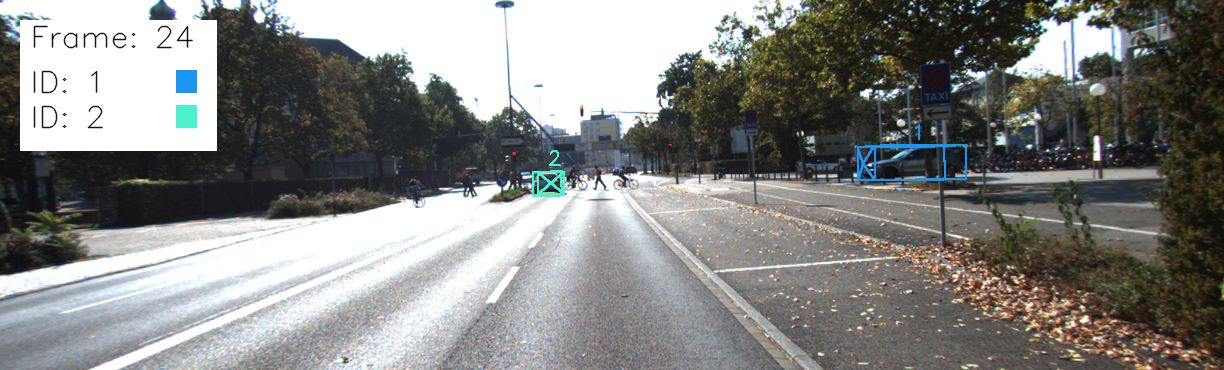}
    \end{minipage}%
     \begin{minipage}[b]{0.33\linewidth}
        \centering
        \includegraphics[width=\linewidth,trim={0cm 3.8cm 0cm 0.5cm},clip]{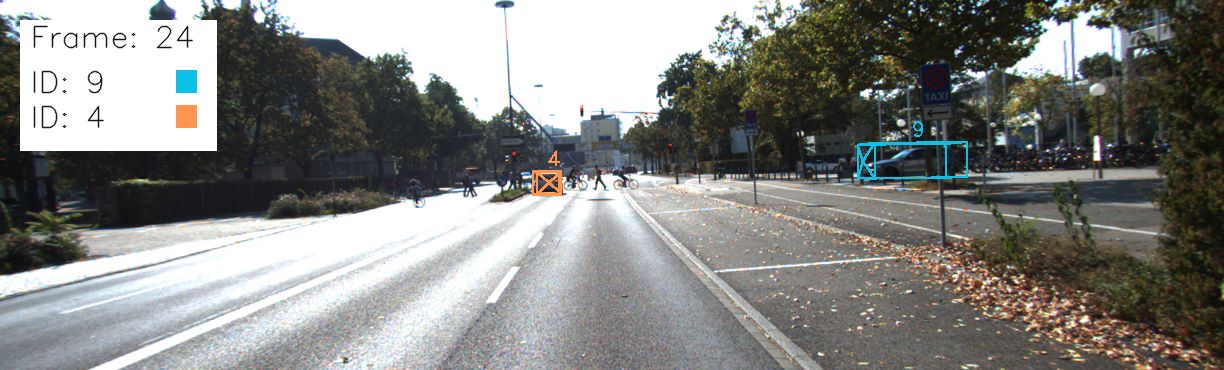}
    \end{minipage}%
     \begin{minipage}[b]{0.33\linewidth}
        \centering
        \includegraphics[width=\linewidth,trim={0cm 3.8cm 0cm 0.5cm},clip]{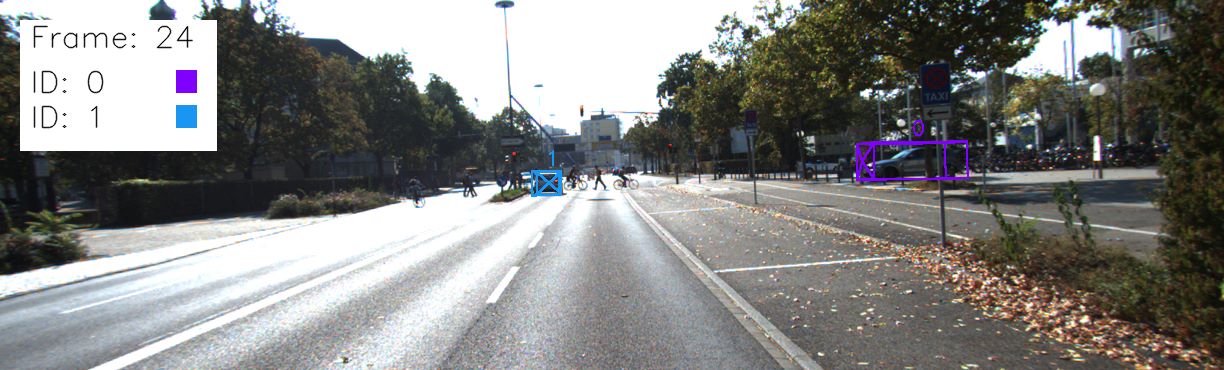}
    \end{minipage}\\[0.01cm]
    \begin{minipage}[b]{0.33\linewidth}
        \centering
        \includegraphics[width=\linewidth,trim={0cm 3.8cm 0cm 0.5cm},clip]{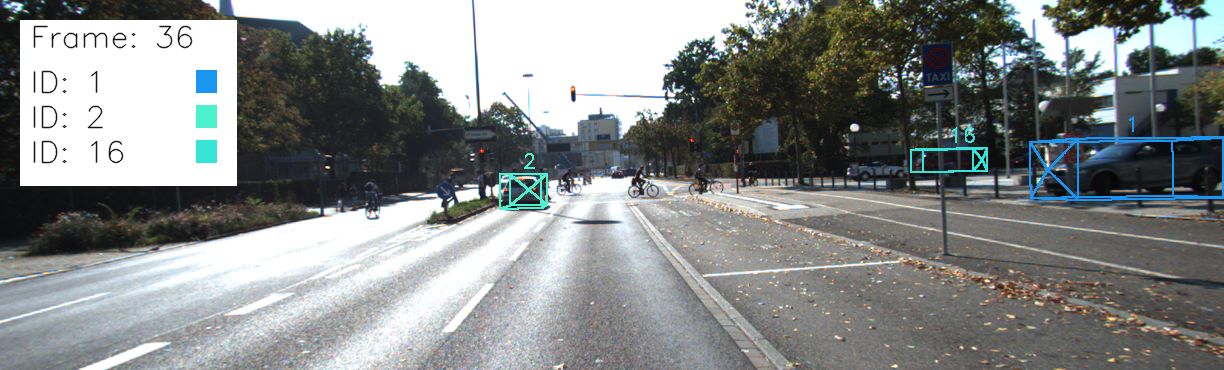}
    \end{minipage}%
     \begin{minipage}[b]{0.33\linewidth}
        \centering
        \includegraphics[width=\linewidth,trim={0cm 3.8cm 0cm 0.5cm},clip]{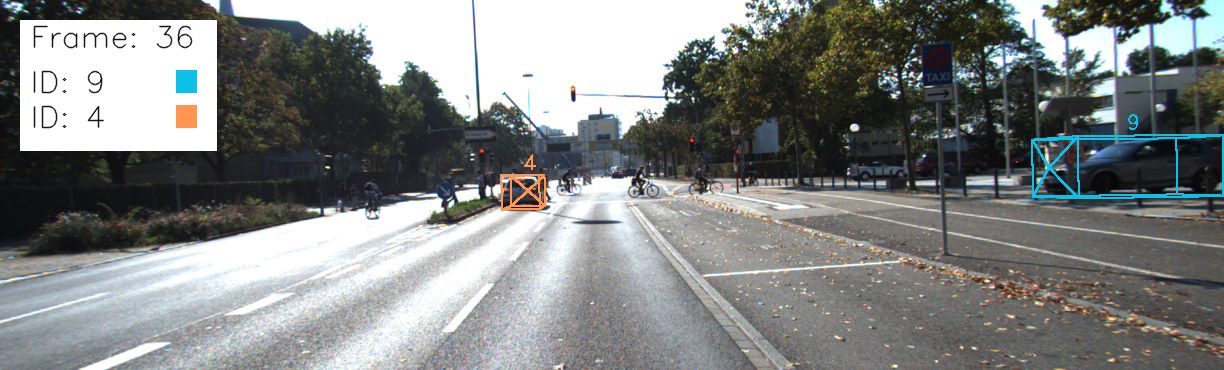}
    \end{minipage}%
     \begin{minipage}[b]{0.33\linewidth}
        \centering
        \includegraphics[width=\linewidth,trim={0cm 3.8cm 0cm 0.5cm},clip]{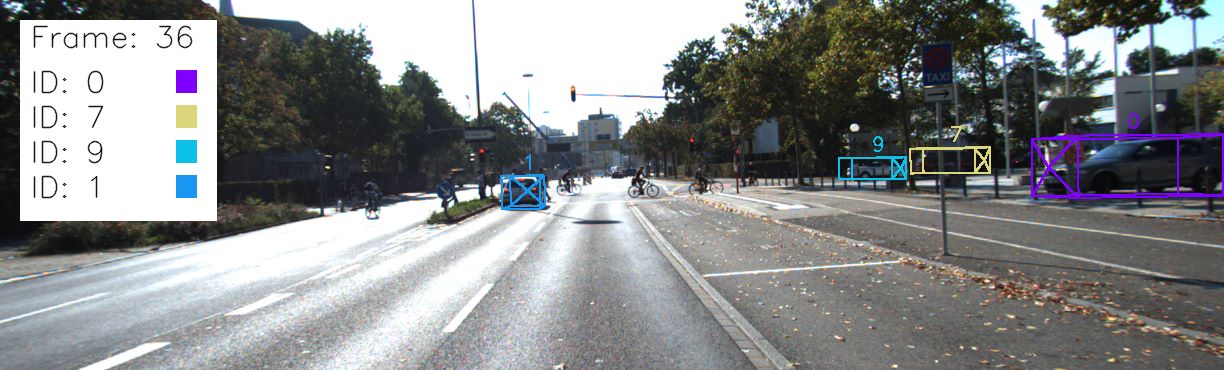}
    \end{minipage}\\[0.01cm]
    \begin{minipage}[b]{0.33\linewidth}
        \centering
        \includegraphics[width=\linewidth,trim={0cm 2cm 0cm 2.0cm},clip]{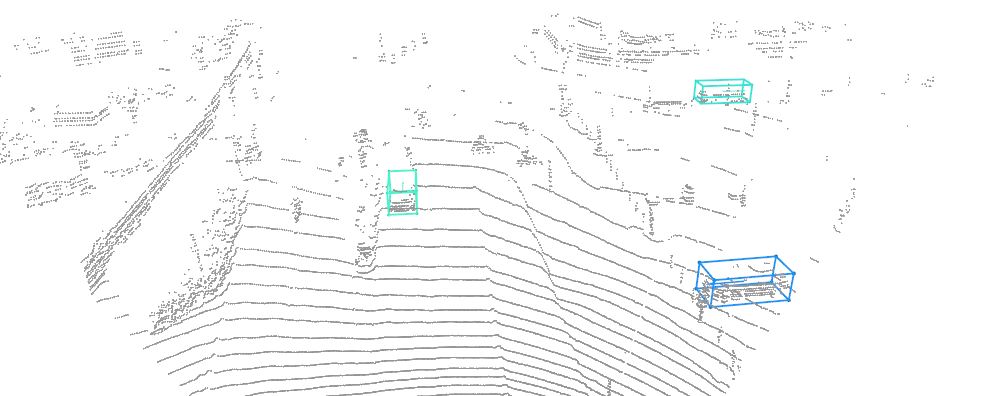}
        {$\text{(a) Ground Truth}$}
    \end{minipage}%
         \begin{minipage}[b]{0.33\linewidth}
        \centering
        \includegraphics[width=\linewidth,trim={0cm 2cm 0cm 2.0cm},clip]{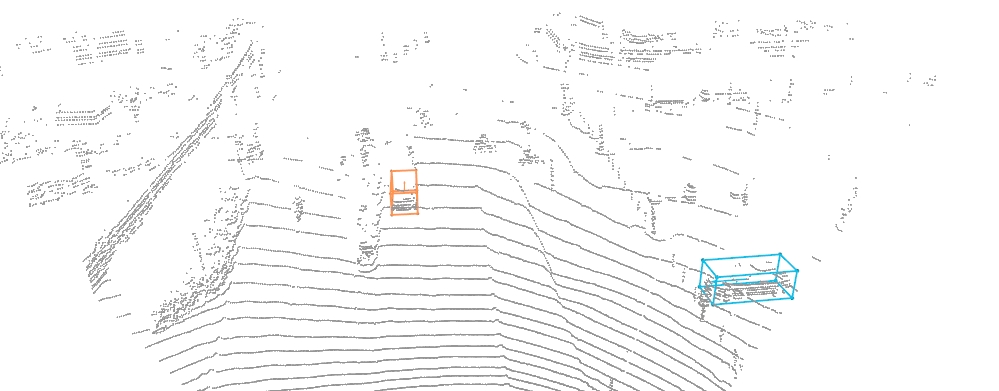}
        {$\text{ (b) UG3DMOT}^*$ \cite{he20243d}}
    \end{minipage}%
     \begin{minipage}[b]{0.33\linewidth}
        \centering
        \includegraphics[width=\linewidth,trim={0cm 2cm 0cm 2.0cm},clip]{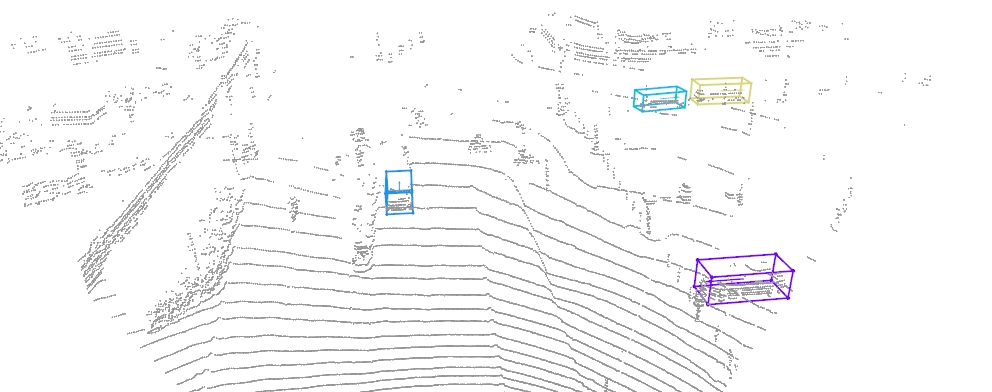}
        {$\text{(c) HybridTrack}^*$}
    \end{minipage}\\[0.01cm]
    \caption{Qualitative results comparison between ground truth, UG3DMOT and HybridTrack on sequence 15 of the validation set. $^*$ uses CasA \cite{wu2022casa} detector.}
\label{fig:2x6_grid_1}
\vspace{-0.5cm}

\end{figure*}

{\textbf{\textit{Comparison with state-of-the-art methods}}. The performance of our proposed method against the existing methods is shown in Tab. \ref{tab:comparison_of_methods} and \ref{tab:clear_metrics}. Our approach achieves near SOTA performance  with a HOTA score of 82.72\% and a detector accuracy of 79.28\%, outperforming the majority of methods that utilize camera-LiDAR data fusion, while our method relies solely on LiDAR data. BiTrack \cite{huang2024bitrack}, while slightly outperforming in certain metrics, benefits from an intensive post-processing pipeline, including single- and multi-trajectory re-optimization, merging, and smoothing, which significantly contributes to its final scores. Moreover, our one-step association approach achieves state-of-the-art association accuracy of 86.92\%, surpassing two-step methods such as MCTrack \cite{wang2024mctrack}, which achieves 86.85\%. It is worth mentioning that in MCTrack, motion models and noise covariances for each individual state variable (e.g., position, size, yaw) and agent types are manually fine-tuned. One can note that in both detection and association, our method prioritizes precision over recall. This approach minimizes false detection and association rates at the cost of introducing missed detections or associations. Additionally, although LocA heavily relies on the performance of the pre-trained 3D object detector, HybridTrack demonstrates strong spatial precision when vehicles are detected or when detections are generated using our LKF. } HybridTrack achieves superior MOTA (91.05\% with Virconv and 90.16\% with CasA) and Mostly Tracked (MT) rate (85.39\% and 84.61\%) compared to UGD3MOT (87.98\%, 79.08\%), as shown in Tab. \ref{tab:clear_metrics}. This highlights our tracker’s ability to maintain object continuity over trajectories, even with occasional misdetections or identity swaps. High MT and low PT rates emphasize long-term continuity, crucial for real-world applications like ADAS. In Fig. \ref{fig:2x6_grid_1}, we present comparative visualizations demonstrating our method's effectiveness in handling long-term occlusions and distant vehicles, two significant challenges in autonomous driving scenarios. The results show several key improvements over UG3DMOT. For example, when vehicle ID=0 (shown in a,b,c) becomes occluded for more than ten frames, UG3DMOT incorrectly initiates a new trajectory (ID=9 in b). In contrast, our method successfully re-establishes the correct association when the vehicle reappears in frame 24. Additionally, our method effectively tracks distant vehicles (IDs=7,9 in c) despite occlusions. While UG3DMOT doesn't track these vehicles due to noisy initial detections, our method maintains consistent tracking, with the scaling mechanism during early detection stages.
\section{Ablation Study}
\textbf{\textit{3D Detector impact}}. Tab. \ref{tab:detector_perf} demonstrates that the HybridTrack tracker’s performance (HOTA) strongly correlates with the underlying 3D detector’s accuracy, as higher-performing detectors (e.g., VirConv \cite{wu2023virtual}, CasA \cite{wu2022casa}) yield superior tracking results.  However, one can note that the tracker is robust to detector performance drops, with more minor losses in HOTA compared to the drop in detection AP. For instance, while Second IoU has a 13.54\% lower Moderate AP than VirConv-S, its HOTA decreases only by 7.77\%.
\begin{table}[h!]
\caption{Ablation Study on 3D Detector Adaptability for KITTI and Corresponding Detection Performance. The results are reported in \%.\label{tab:detector_perf}}
\resizebox{\columnwidth}{!}{%
\centering
\begin{tabular}{c|cc|c}
\toprule
\multirow{1}{*}{\textbf{3D Detector Choice}} & \multicolumn{1}{c}{{AP (IoU=0.7) Moderate} } & \multicolumn{1}{c|}{AP (IoU=0.7) Easy} & \multirow{1}{*}{\textbf{HOTA} $\uparrow$} \\
\midrule
Second IoU \cite{yan2018second} & 73.66  & 83.13  & {78.44}  \\
PV-RCNN \cite{shi2020pv} & 81.43  & 90.25  & {80.14}  \\
CasA \cite{wu2022casa} & 83.06  & 91.58  & {83.59}  \\
VirConv-S \cite{wu2023virtual} & 87.20  & 92.48  & \textbf{86.35}  \\
\bottomrule
\end{tabular}}
\vspace{-0.5cm}
\end{table}

\begin{figure}[h]
  \centering
 \includegraphics[width=\columnwidth,trim={0cm 0.2cm 0cm 1.0cm},clip]{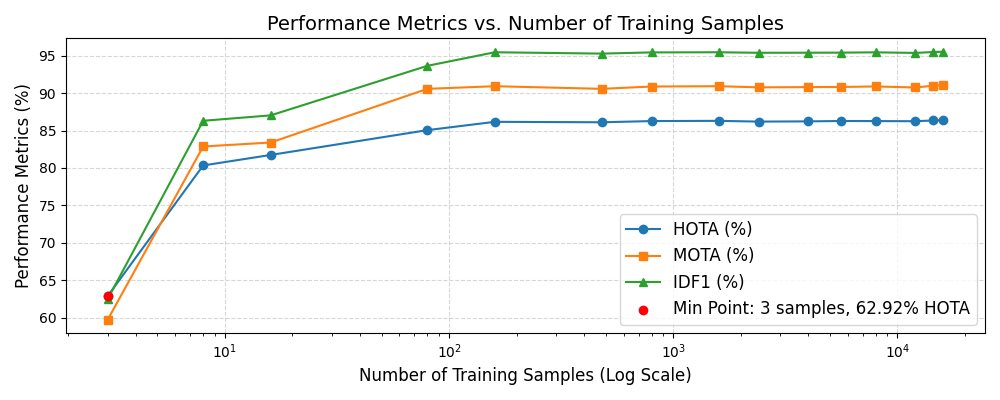}
  \caption{Performance Metrics vs. Training dataset size.}
  \label{fig:ratio_dataset}
\vspace{-0.1cm}
\end{figure}

\textbf{\textit{Model Performance vs. Dataset Size}}. One of the main advantages of traditional filtering approaches against deep-learning-based methods is that these model-based methods do not need any data to be trained on. In this experiment, we show that our proposed method, HybridTrack, doesn't require a lot of data to be competitive with non-data-driven methods. As illustrated in Fig. \ref{fig:ratio_dataset}, our model demonstrates strong data efficiency, achieving 81\% HOTA on tracking on the validation set when trained with only 16 sequences (320 timesteps), compared to 86.355\% HOTA when trained on 16,000 sequences. Also, one can see that, with a dataset containing only three vehicle trajectories, HybridTrack can already reach 62.92\%, as illustrated by the red dot. Moreover, unlike non-deep-learning methods, which tend logically to plateau quickly regardless of the dataset size, our approach is scalable. It can continue to learn from additional data, enabling it to capture specific edge cases and consistently improve its object-tracking performance. One potential use case for our method could be handling accident scenarios, cases that are often lacking in publicly available datasets like KITTI. Unlike classical models that struggle with the unpredictability of accident trajectories, our data-driven approach can learn from accident data and would be able to adapt to these challenging situations, enabling more robust and reliable tracking in real-world applications.
\vspace{-0.1cm}

\textbf{\textit{Data Association Cost Function Choice}}. Tab. \ref{tab:tab_cost_map} evaluates cost map choices and their impact on tracking performance, highlighting a trade-off between accuracy and computational efficiency. The 3D-CIoU cost map achieves the best tracking performance (HOTA: 86.355\%) but is computationally expensive, with the lowest FPS (98.91). In contrast, the simple Euclidian distance-based cost map is the fastest (FPS: 112.83) while maintaining solid performance (HOTA:  85.864\%). Combining the L2 error and the size error slightly improves the accuracy. The 2D CIoU method performs worst across metrics, highlighting the importance of 3D domain association for better tracking.
\vspace{-0.2cm}

\begin{table}[h!]
\caption{Ablation Study on Cost Map Choices and Metrics Evaluation. The results are reported in \%.\label{tab:tab_cost_map}}
\resizebox{\columnwidth}{!}{%
\centering
\begin{tabular}{cccc|cccc}
\toprule
\textbf{2D CIoU} & \textbf{L2 Error} & \textbf{Size Error} & \textbf{3D CIoU} $\uparrow$  & \textbf{HOTA} $\uparrow$  & \textbf{MOTA} $\uparrow$  & \textbf{IDFP} $\uparrow$  & \textbf{FPS} $\uparrow$  \\
\midrule
\checkmark & - & -& - & 82.052  & 86.574  & 88.742  & 44.81\\
- & \checkmark & -           & - &  85.964  & 90.345   & 94.988  & \textbf{112.83} \\
- & \checkmark    & \checkmark & - & 86.037  & 90.500  & 95.084  & 111.60\\
-           & - & - & \checkmark &  \textbf{86.355 } & \textbf{91.061 }  & \textbf{95.512 } & 98.91 \\
\bottomrule
\end{tabular}}
\vspace{-0.1cm}

\end{table}

{\textbf{\textit{Ablation Study on Learning-based Components}}. We conduct an ablation study to evaluate the impact of our learnable prediction and update steps. As shown in Table~\ref{tab:tab_learned}, learning only the update step fails to converge effectively, indicating its dependence on an accurate predicted state. In contrast, learning the prediction step alone already yields strong performance, demonstrating its important role in modeling motion dynamics. Jointly learning both steps achieves the best performance, highlighting the benefit of end-to-end optimization. These results confirm that combining L-PS and L-US significantly improves tracking accuracy compared to a standard Kalman filter with a constant-acceleration model.
}
\begin{table}[h!]
\caption{{Ablation study on learning-based components. 'L-PS' refers to the Learned Prediction Step, 'L-US' to the Learned Update Step. Results are reported in \%.\label{tab:tab_learned}}}
\centering
\begin{tabular}{lcccc}
\toprule
\textbf{Configuration} & \textbf{Baseline} & \textbf{L-US Only} & \textbf{L-PS Only} & \textbf{L-PS + L-US} \\
\midrule
\textbf{HOTA} $\uparrow$ & 83.927 & 54.320 & 85.276 & \textbf{86.355} \\
\bottomrule
\end{tabular}
\end{table}



\begin{figure}[h]
  \centering
  \begin{subfigure}[b]{0.48\columnwidth}
    \centering
    \includegraphics[width=\columnwidth, trim={0.5cm 3cm 0.5cm 0.5cm}, clip]{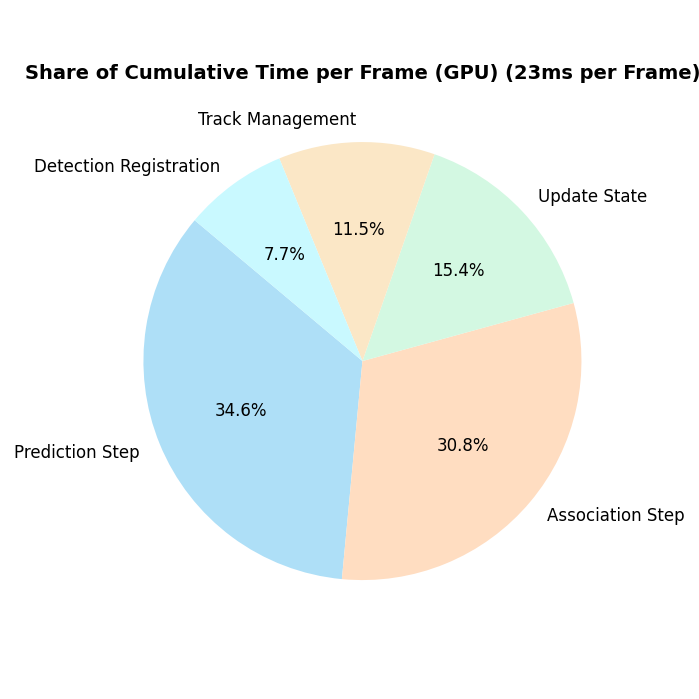}
    \label{fig:cumtime}
  \end{subfigure}
  \hfill
  \begin{subfigure}[b]{0.49\columnwidth}
    \centering
    \includegraphics[width=\columnwidth, trim={0.5cm 2.8cm 0.5cm 0.5cm}, clip]{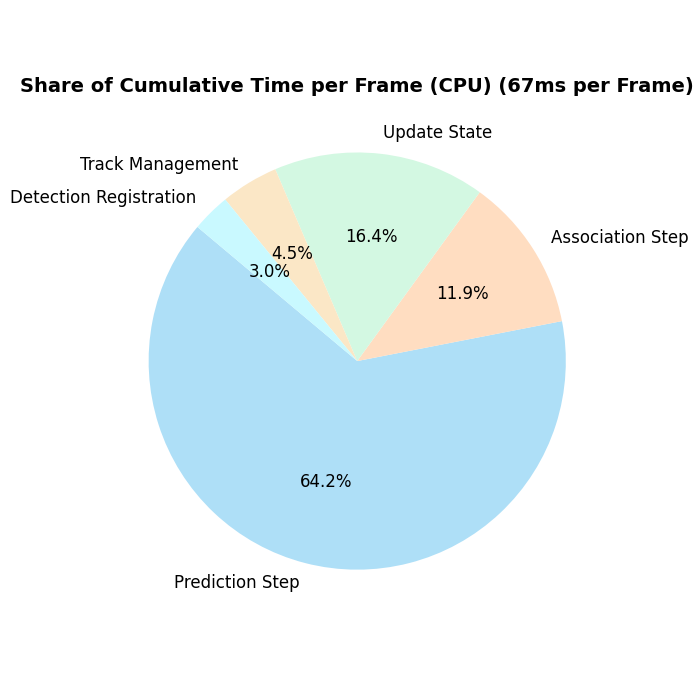}
    \label{fig:another}
  \end{subfigure}
  \caption{{Share of Cumulative Execution Time for Sequence 1 of the validation set.}}
  \label{fig:compute}
\vspace{-0.9cm}
\end{figure}

{\textbf{\textit{Computational Cost}}. HybridTrack achieves high efficiency during inference on a Titan X NVIDIA GPU and an Intel(R) Xeon(R) CPU E5-2630 v4 @ 2.20GHz CPU, respectively. Our method achieves an average of 98.91 FPS across all test sequences on a GPU device. On the other hand, while our method is optimised for GPU devices, it remains reasonably efficient on CPU devices, on average 20.80 frames can be processed per second. In Fig. \ref{fig:compute}, we visualize the execution time usage percentage of each step in HybridTrack. The primary computational bottleneck remains in the learning-based motion prediction and association steps. }
\vspace{-0.5cm}

\section{Conclusion}
\label{sec:page}
We introduce HybridTrack, a 3D MOT framework that fuses Kalman filtering with deep learning for robust, efficient, and scalable tracking. HybridTrack eliminates manual tuning by integrating learnable motion models and noise covariances, adapting to diverse traffic scenarios. It achieves state-of-the-art performance and demonstrates strong data efficiency and scalability. Our approach is well-suited for real-world traffic applications, offering reliable tracking even in challenging use cases such as long-term occlusion and distant vehicles. 
\vspace{-0.4cm}


%



%

\bibliographystyle{IEEEtran}
\bibliography{IEEEabrv, root.bib}

%




\end{document}